\documentclass{article}

\usepackage[preprint]{corl_2024} 
\usepackage{graphicx}
\usepackage{pgfplots}
\usepackage{amsfonts,amssymb,amsmath,bm, mathabx}
\usepackage[ruled,vlined]{algorithm2e}
\usepackage{multirow}

\usepackage{subfig}
\usepackage{float}
\usepackage{svg}
\usepackage{wrapfig}
\usepackage{algorithmic}
\usepackage{eqparbox}

\usepackage{cleveref}
\usepackage{musicography}

\usepackage{tabularray}

\usepackage[automake]{glossaries}
\newacronym{ifo}{IfO}{Imitation Learning from Observations}
\newacronym{bc}{BC}{Behavioral Cloning}
\newacronym{irl}{IRL}{Inverse Reinforcement Learning}
\newacronym{rl}{RL}{Reinforcement Learning}
\newacronym{ik}{IK}{Inverse Kinematics}
\newacronym{mdp}{MDP}{Markov Decision Process}
\newacronym{ppo}{PPO}{Proximal Policy Optimization}
\newacronym{sdf}{SDF}{Signed Distance Field}
\makeglossaries

\def\1{\bm{1}}

\def\RR{\mathbb{R}}

\def\vtheta{{\bm{\theta}}}
\def\va{{\bm{a}}}

\def\vg{{\bm{g}}}

\def\vq{{\bm{q}}}

\def\vs{{\bm{s}}}

\def\vx{{\bm{x}}}

\def\vz{{\bm{z}}}

\def\vtheta{{\boldsymbol{\theta}}}
\def\vtau{{\boldsymbol{\tau}}}

\DeclareMathAlphabet{\mathsfit}{\encodingdefault}{\sfdefault}{m}{sl}
\SetMathAlphabet{\mathsfit}{bold}{\encodingdefault}{\sfdefault}{bx}{n}

\def\gD{{\mathcal{D}}}

\usepackage{hyperref}
\hypersetup{
 colorlinks=teal,
 filecolor=teal,
 citecolor=teal,      
 urlcolor=teal,
 }

\title{PianoMime: Learning a Generalist, Dexterous Piano Player from Internet Demonstrations}
\title{PianoMime: Learning a Generalist, Dexterous Piano Player from Internet Demonstrations}

\author{
  Cheng Qian\\
  TU Munich\\ 
  \And
  Julen Urain \\
  TU Darmstadt \\
  \And
  Kevin Zakka\\
  UC Berkeley\\
  \And
  Jan Peters\\
  TU Darmstadt\\
}

\begin{document}
\maketitle


\begin{abstract}
In this work, we introduce PianoMime, a framework for training a piano-playing agent using internet demonstrations.
The internet is a promising source of large-scale demonstrations for training our robot agents. 
In particular, for the case of piano-playing, Youtube is full of videos of professional pianists playing a wide myriad of songs.
In our work, we leverage these demonstrations to learn a generalist piano-playing agent capable of playing any arbitrary song.
Our framework is divided into three parts: a data preparation phase to extract the informative features from the Youtube videos,
a policy learning phase to train song-specific expert policies from the demonstrations and a policy distillation phase to distil the policies into a single generalist agent.
We explore different policy designs to represent the agent and evaluate the influence of the amount of training data on the generalization capability of the agent to novel songs not available in the dataset.
We show that we are able to learn a policy with up to 56\% F1 score on unseen songs. Project website: \href{https://pianomime.github.io/}{https://pianomime.github.io/}
\end{abstract}



\section{Introduction}
\label{sec:introduction}

The Internet is a promising source of large-scale data for training generalist robot agents. If properly exploited, it is full of demonstrations (video, text, audio) of humans solving an infinite amount of tasks~\cite{volske2017tl,  fan2022minedojo, grauman2022ego4d} that could inform our robot agents on how to behave.
However, learning from these databases is challenging for several reasons.
First, unlike teleoperation demonstrations, video data does not specify the actions applied by the robot, usually requiring the use of reinforcement learning to induce the robot actions~\cite{torabi2019recent,fan2022minedojo, peng2018deepmimic}.
Second, videos typically show a human performing the task, while the learned policy is deployed on a robot. This often requires to re-target the human motion to the robot body~\cite{peng2018deepmimic, garcia2020physics, peng2020learning}. 
Finally, as pointed in~\cite{fan2022minedojo}, if we aim to learn a generalist agent, we must select a task for which large-scale databases are available and that allows an unlimited variety of open-ended goals.


From opening doors~\cite{garcia2020physics} to rope manipulation~\cite{nair2017combining} or pick and place tasks~\cite{shao2021concept2robot, ma2022vip}, previous works have successfully taught robot manipulation skills through observations.
However, these approaches have been limited to low dexterity in the robots or to a small variety of goals.

In this work, we focus on the task of \textbf{learning a generalist piano player from Internet demonstrations}.
Piano-playing is a highly dexterous open-ended task~\cite{zakka2023robopianist}.
Given two multi-fingered robot hands and a desired song, the goal of a piano-playing agent is to press the correct keys and only the correct keys at the proper timing.
Moreover, the task can be conditioned on arbitrary songs, allowing for a large, and high-dimensional goal conditioning.
\\
Additionally, the Internet is full of videos of professional piano players performing a wide myriad of songs. Interestingly, these piano players often record themselves from a top-view allowing an easy observation of the demonstrations. Additionally, they usually share the MIDI files of the song they play, facilitating the extraction of relevant information.

\begin{figure}
    \centering
    \includegraphics[width=.99\textwidth]{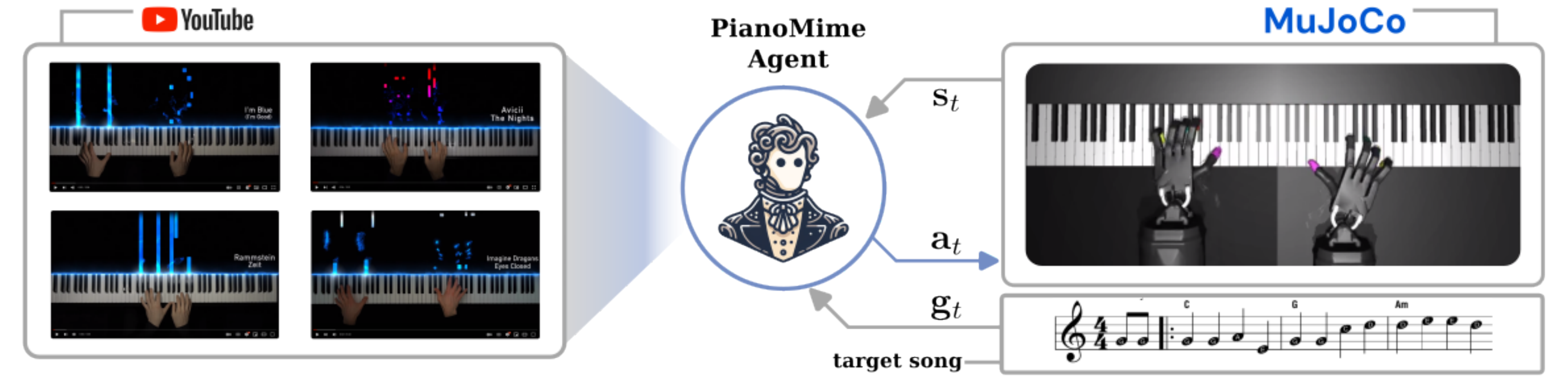}
    \caption{The goal of this work is to train a generalist piano-playing agent (PianoMime) from Youtube videos. We collect a set of videos and accompanying MIDI files and train a single agent to play any song, combining reinforcement learning and behavioral cloning.}
    \label{fig:main_figure}
    \vspace{-.3cm}
\end{figure}

To learn a generalist piano-playing agent from internet data, we introduce \textbf{PianoMime}, a framework to train a single policy capable of playing any song (See \Cref{fig:main_figure}). 
In its essence, the PianoMime agent is a goal-conditioned policy that generates configuration space actions given the desired song to be played. At each timestep, the agent receives as goal input a trajectory of the keys to be pressed. Then, the policy generates a trajectory of actions and executes them in chunk. 
\\
\textbf{To learn the agent,} we combine both reinforcement learning with imitation learning. 
We train individual song-specific expert policies by using reinforcement learning in conjunction with Youtube demonstrations and we distill all the expert policies into a single generalist behavioral cloning policy.
\\
\textbf{To represent the agent,} we perform ablations of different architectural design strategies to model the behavioral cloning policy.
We investigate the benefit of incorporating representation learning to enhance the geometric information of the goal input.
Additionally, we explore the effectiveness of a hierarchical policy that combines a high-level policy generating fingertip trajectories with a learned \textit{cross-domain inverse dynamics model} generating joint-space actions.
We show that the learned agent is able to play arbitrary songs not included in the training dataset with around 56\% F1-score.

In summary, \textbf{the main contribution} of this work is a framework for training a generalist piano-playing agent using Internet demonstration data.
To achieve this goal, we:
\begin{itemize}
\item Introduce a method to learn policies from the internet demonstrations by decoupling the human motion information from the task-related information.
\item Present a reinforcement learning approach that combines residual policy learning strategies~\cite{silver2018residual, johannink2019residual} with style reward-based strategies~\cite{peng2018deepmimic}.
\item  Explore different policy architecture designs, introducing novel strategies to learn geometrically consistent latent features and conducting ablations on different architectural designs. 
\end{itemize}
Finally, we are releasing the dataset and the trained models as a benchmark for testing internet-data-driven dexterous manipulation.

\section{Related Work}
\label{sec:related_work}
\textbf{Robotic Piano Playing}
Several studies have investigated the development of robots capable of playing the piano. In \cite{scholz2019}, multiple-targets \gls{ik} and offline trajectory planning are utilized to position the fingers above the intended keys. In \cite{xu2022towards}, a \gls{rl} agent is trained to control a single Allegro hand to play the piano using tactile sensor feedback. However, the piano pieces used in these studies are relatively simple. Subsequently, in \cite{zakka2023robopianist}, an RL agent is trained to control two Shadow Hands to play complex piano pieces by designing a reward function comprising a fingering reward, a task reward, and an energy reward.
In contrast with previous approaches, our approach exploits Youtube piano-playing videos, enabling faster training and more accurate and human-like robot behavior.


\textbf{Motion Retargeting and Reinforcement Learning}
Our work shares similarities with motion retargeting~\cite{geijtenbeek2013flexible}, specifically with those works that combine motion retargeting with RL to learn control policies~\cite{chentanez2018physics, liu2018learning, peng2018deepmimic, peng2021amp, garcia2020physics}.
Given a mocap demonstration, it has been common to exploit the demonstration rather as a reward function~\cite{peng2018deepmimic, peng2021amp} or as a nominal behavior for residual policy learning~\cite{liu2018learning, garcia2020physics}.
In our work, we not only extract the mocap information, but also task-related information (piano states) allowing the agent to balance between mimicking the demonstrations and solving the task.

\section{Method}
\label{sec:pianomime}
The PianoMime framework is composed of three phases: data preparation, policy learning, and policy distillation.
\\
In the \textbf{data preparation phase}, given the raw video demonstration, we extract the informative signals needed to train the policies. 
Specifically, we extract fingertip trajectories and a MIDI file that informs the piano state at every time instant.
\\
In the \textbf{policy learning phase}, we train song-specific policies via \gls{rl}.
This step is essential for generating the robot actions that are missing in the demonstrations.
The policy is trained with two reward functions: a style reward and a task reward. 
The style reward aims to match the robot's finger movements with those of the human in the demonstrations to preserve the human style, while the task reward encourages the robot to press the correct keys at the proper timing.
\\
In the \textbf{policy distillation phase}, we train a single behavioral cloning policy to mimic all the song-specific policies.
The goal of this phase is to train a single generalist policy capable of playing any song.
We explore different policy designs and the representation learning of goals to improve the generalization capability of the policy.


\subsection{Data preparation: From raw data to human and piano state trajectories}
We generate the training dataset by web scraping. We download YouTube videos of professional piano artists playing various songs.
We particularly choose YouTube channels that also upload MIDI files of the played songs. The MIDI files represent trajectories of the piano state (pressed/unpressed keys) throughout the song.
We use the video to extract the motion of human pianists and the MIDI file to inform about the goal state of piano during the execution of the song.
\\
We select the fingertip position as the essential signal to mimic with the robot hand.
While several dexterous tasks might require the use of the palm (e.g. grasping a bottle), we consider mimicking the fingertip motion to be sufficient for the task of piano playing. This will also reduce the constraints applied to the robot, allowing it to adapt its embodiment more freely.
\\
To extract the fingertip motion from videos, we use MediaPipe~\cite{lugaresi2019mediapipe}, an open-source framework for perception. Given a frame from the demonstration videos, MediaPipe outputs the skeleton of the hand.
We find that the classical top-view recording in piano-playing YouTube videos is highly beneficial for obtaining an accurate estimate of the fingertip positions.
\\
Notice that given the videos are RGB, we lack depth signal. Therefore, we predict the 3D fingertip positions based on the piano state. The detailed procedure is explained in \Cref{app:retarget}.

\subsection{Policy learning: generating robot actions from observations}
Through the data preparation phase, we extract two trajectories: a human fingertip trajectory $\vtau_x$ and a piano state trajectory $\vtau_{\musEighth}$.
The human fingertip trajectory $\vtau_{\vx}:(\vx_1,\dots,\vx_T)$ is a $T$-step trajectory of two hands' 3D fingertip positions  $\vx \in \RR^{3\times10}$ (10 fingers). The piano state trajectory $\vtau_{\musEighth}: (\musEighth_{1}, \dots, \musEighth_{T})$ is a $T$-step trajectory of piano states $\musEighth \in \mathbb{B}^{88}$, represented with an 88-dimensional binary variable representing which keys should be pressed.

Given the \textsc{RoboPianist}~\cite{zakka2023robopianist} environment, \textbf{our goal} is to learn a goal-conditioned policy $\pi_{\vtheta}$ that plays the song defined by $\vtau_{\musEighth}$ while matching the fingertip motion given by $\vtau_{\vx}$. 
Notice that satisfying both objectives jointly might be impossible. Tracking perfectly the fingertip trajectory $\vtau_{\vx}$ might not necessarily lead to playing the song correctly.
Although both trajectories are collected from the same source, errors in hand tracking and embodiment mismatches might lead to deviations, resulting in poor song performance. 
Thus, we propose using $\vtau_{\vx}$ as a style guiding behavior.

Similarly to \cite{zakka2023robopianist}, we formulate the piano playing as an \textbf{\gls{mdp}} with the horizon of the episode $H$, being the duration of the song to be played.
The state observation is defined by the robot's proprioception $\vs$ and the goal state $\vg_t$. The goal state $\vg_t$ at time $t$ informs the desired piano key configurations $\musEighth$ in the future $\vg_t = (\musEighth_{t+1}, \dots, \musEighth_{t+L})$, with $L$ being the lookahead horizon. 
As claimed in \cite{zakka2023robopianist}, to successfully learn how to play, the agent needs to be aware of several steps into the future to plan its actions.
The action $\va$ is defined as the desired configuration for both hands $\vq\in\RR^{23\times2 + 1}$, each with 23 joint angles and one dimension for the sustain pedal.

We propose solving the reinforcement learning problem by combining residual policy learning~\cite{silver2018residual, johannink2019residual, garcia2020physics} and style mimicking rewards~\cite{peng2018deepmimic, peng2021amp}. 
\\
\textbf{Residual policy architecture.}
Given the fingertip trajectory $\vtau_{\vx}$, we solve an \gls{ik}~\cite{pink2024} problem to obtain a trajectory of desired joint angles $\vtau_{\vq}^{\text{ik}}:(\vq_0^\text{ik},\dots,\vq_T^\text{ik})$ for the robot hands.
Then, we represent the policy $\pi_{\vtheta}(\va|\vs,\vg_t) = \pi_{\vtheta}^{r}(\va|\vs,\vg_t) + \vq_{t+1}^\text{ik}$ as a combination of a nominal behavior (given by the \gls{ik} solution) and a residual policy $\pi_{\vtheta}^{r}$. 
Given the goal state at time $t$, the nominal behavior is defined as the next desired joint angle $\vq_{t+1}^{\text{ik}}$. 
We then only learn the residual term around the nominal behavior. 
In practice, we initialize the robot at $\vq_0^{\text{ik}}$ and roll both the goal state and the nominal behavior with a sliding window along $\vtau_{\musEighth}$ and $\vtau_{\vq}^{\text{ik}}$ respectively.
\\
\textbf{Style-mimicking reward.}
We also integrate a style-mimicking reward to preserve the human style in the trained robot actions.
The reward function $r = r_{\musEighth} + r_{\vx}$ is composed of a task reward $r_{\musEighth}$ and a style-mimicking reward $r_{\vx}$. While the task reward $r_{\musEighth}$ encourages the agent to press the correct keys, the style reward $r_{\vx}$ encourages the agent to move its fingertips similar to the demonstration $\vtau_{\vx}$. 
We provide further details in \Cref{app:mdp}.

\subsection{Policy distillation: learning a generalist piano-playing agent}
\label{sec: dis}
Through the policy learning phase, we train song-specific expert policies from which we roll out state and action trajectories $\tau_{\vs}:(\vs_0,\dots,\vs_T)$ and $\tau_{\vq}:(\vq_0,\dots,\vq_T)$. Then, we generate a dataset $\gD:(\tau_{\vs}^i, \tau_{\vq}^i,\tau_{\vx}^i,\tau_{\musEighth}^i)_{i=1}^N$ with $N$ being the number of learned songs.
Given the dataset $\gD$, we apply \gls{bc} to learn a single generalist piano-playing agent $\pi_{\vtheta}(\vq_{t:t+L}, \vx_{t:t+L}|\vs_t , \musEighth_{t:t+L})$ that outputs configuration-space actions $\vq_{t:t+L}$ and fingertip motion $\vx_{t:t+L}$ conditioned on the current state $\vs_t$ and the future desired piano state $\musEighth_{t:t+L}$.
\begin{figure}
    \centering
    \includegraphics[width=.99999\textwidth]{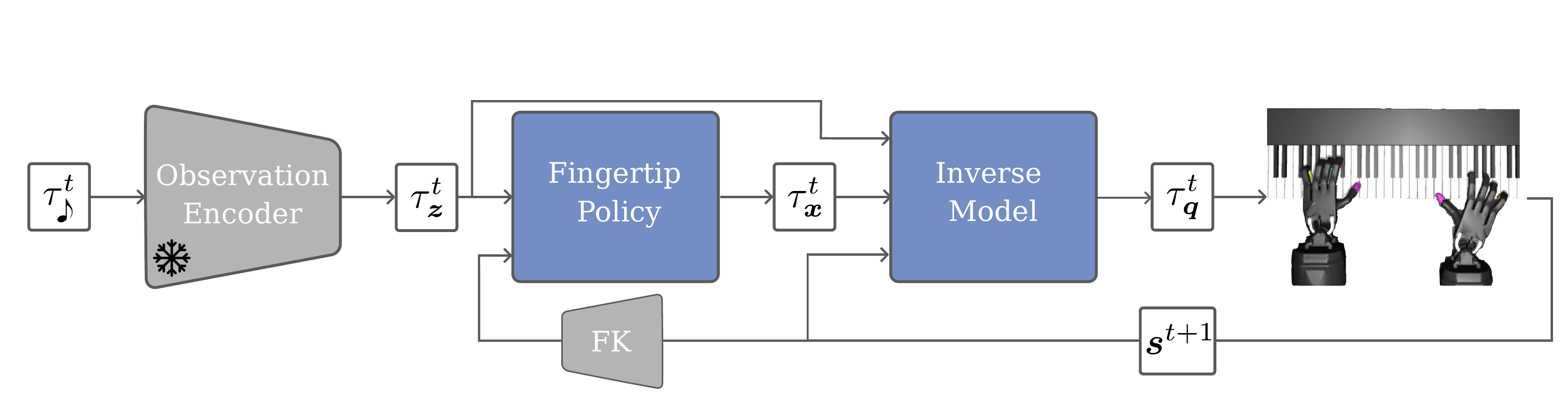}
    \caption{Proposed distillation policy architecture. Given a L steps window of a target song $\tau^t_{\musEighth} :(\musEighth_{t:t+L})$ at time $t$, a latent representation $\tau_\vz^t$ is computed given a pre-trained observation encoder. Then, the policy is decoupled between a high-level fingertip predictor that generates a trajectory of fingertip positions $\tau_\vx^t$ and a low-level inverse dynamics model that generates a trajectory of target joint position $\tau_{\vq}^t$.}
    \label{fig:pianomime_framework}
    \vspace{-0.5cm}
\end{figure}

We explore different strategies to represent and learn the behavioral cloning policy and improve its generalization capabilities. In particular, we explore (\textbf{1}) representation learning approaches to induce spatially informative features, (\textbf{2}) a hierarchical policy structure for sample-efficient training, and (\textbf{3}) expressive generative models~\cite{chi2023diffusion, shafiullah2022behavior, florence2022implicit} to capture the multimodality of the data.
Also, inspired by current behavioral cloning approaches~\cite{chi2023diffusion, zhao2023learning}, we train policies that output sequences of actions rather than single-step actions and execute them in chunks.
\\
\textbf{Representation Learning.}
We pre-train an observation encoder over the piano state $\musEighth$ to learn spatially consistent latent features. 
We hypothesize that two piano states that are spatially close should lead to latent features that are close. Using these latent features as goal should induce better generalization. 
To obtain the observation encoder, we train an autoencoder with a reconstruction loss over a \gls{sdf} defined on the piano state. Specifically, the encoder compresses the binary vector of the goal into a latent space, while the decoder predicts the \gls{sdf} function value of a randomly sampled query point (the distance between the query point and the closest "on" piano key).
We provide further details in \Cref{app:bc}. 
\\
\textbf{Hierarchical Policy.}
We represent the piano-playing agent with a hierarchical policy. The high-level policy receives a sequence of desired future piano states $\musEighth$ and outputs a trajectory of human fingertip positions $\vx$. Then, a low-level policy takes the fingertip and piano state trajectories as input and outputs a trajectory of desired joint angles $\vq$.
On one hand, while fingertip trajectory data is easily available from the Internet, obtaining low-level joint trajectories requires solving a computationally expensive \gls{rl} problem.
On the other hand, while the high-level mapping ($\musEighth\mapsto\vx$) is complex, which involves fingerings, the low-level mapping ($\vx\mapsto\vq$) is relatively simpler, which addresses a cross-embodiment inverse dynamics problem.
This decoupling allows us to train the more complex high-level mapping on large cheap datasets and the simpler low-level mapping on smaller expensive ones. We visualize the policy in \Cref{fig:pianomime_framework}.
\\
\textbf{Expressive Generative Models.}
Considering that the human demonstration data of piano playing is highly multi-modal, we explore using expressive generative models to better represent this multi-modality. We compare the performance of different deep generative models based policies, e.g., Diffusion Policies \cite{chi2023diffusion} and Behavioral Transformer \cite{shafiullah2022behavior}, as well as a deterministic policy. 


\section{Experimental Results}
\label{sec:result}
We split the experimental evaluation into three parts. 
In the first part, we explore the performance of our proposed framework in learning song-specific policies via \gls{rl}.
In the second part, we perform ablation studies on policy designs for learning a generalist piano-playing agent by distilling the previously learned policies via \gls{bc}.
Finally, in the third part, we explore the influence of the amount of training data on the performance of the test environments.
\\
\textbf{Dataset and Evaluation Metrics}
All experiments are conducted on our collected dataset, which contains the notes and the corresponding demonstration videos and fingertip trajectories of 60 piano songs from a Youtube channel \textbf{PianoX} \footnote{https://www.youtube.com/channel/UCsR6ZEA0AbBhrF-NCeET6vQ}. To standardize the length of each task, each song is divided into several clips, each with a duration of 30 seconds (The dataset contains totally 431 clips, 258K state-action pairs). Furthermore, we choose 12 unseen clips to investigate the generalization capability of the generalist policy. We use the same evaluation metrics from RoboPianist \cite{zakka2023robopianist}, i.e., precision, recall, and F1 score.
\\
\textbf{Simulation Environment}
Our experiment setup utilizes \textsc{RoboPianist} simulation environment \cite{zakka2023robopianist}, conducted in Mujoco physics simulator \cite{6386109}. The agent predicts target joint angles at 20Hz and the targets are converted to torques using PD controllers running at 500Hz. We use the same physical setting as \cite{zakka2023robopianist} with two modifications: 1) The z-axis sliding joints fixed on both forearms are enabled to allow more versatile hand movement. Therefore, the action space of our agent is 47 dimensional (45 dimensional in \cite{zakka2023robopianist}). 2) We increase the proportional gain of the PD controller for the x-axis sliding joints to enable faster horizontal movement, which we find essential for some fast-paced piano songs.
\begin{figure}[t]
  \centering
  \includegraphics[width=\textwidth]{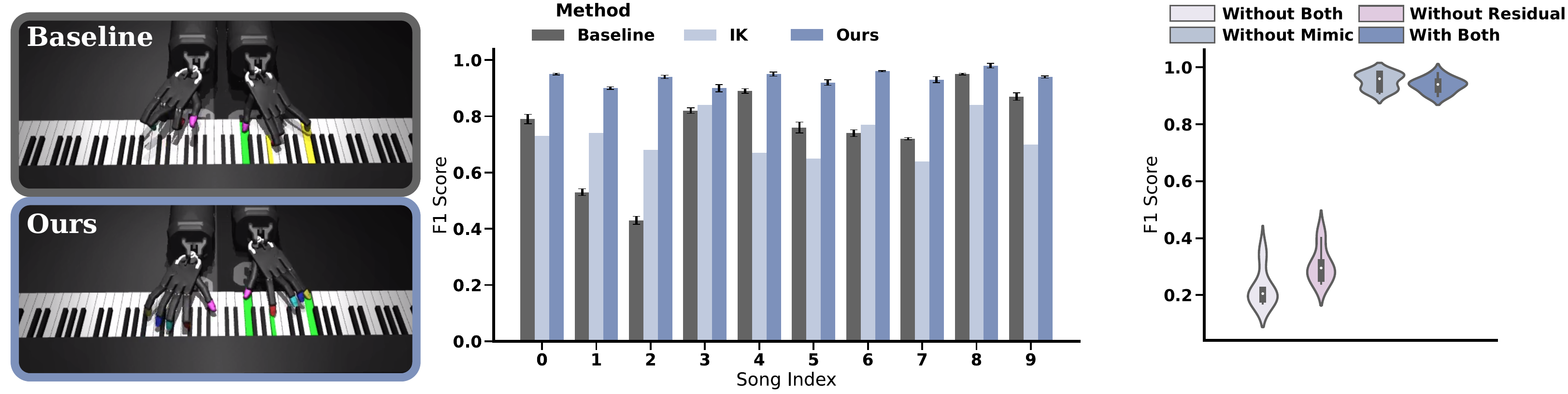}
  \centering
  \caption{Left: Qualitative comparison of hand postures. Middle: The F1 score achieved by three methods for 10 chosen clips; Right: The F1 score achieved by excluding different elements in \gls{rl}.}
  \label{fig:single}
  \vspace{-.7cm}
\end{figure}

\subsection{Evaluation on learning song-specific policies from demonstrations}
In this section, we evaluate the song-specific policy learning and aim to answer the following questions: 
(\textbf{1}) Does integrating human demonstrations with \gls{rl} help in achieving better performance?
(\textbf{2}) What elements of the learning algorithm are the most influential in achieving good performance?


We use Proximal Point Optimization (PPO) \cite{schulman2017proximal}, because we find that it performs the best compared to other \gls{rl} algorithms. We compare our model against two baselines:\\
\textbf{Robopianist~\cite{zakka2023robopianist}} We use the \gls{rl} method introduced in \cite{zakka2023robopianist}.
We maintain the same reward functions as the original work and manually label the fingering from the demonstrations videos to provide the fingering reward.
\begin{wrapfigure}{R}{0.5\textwidth}
    \centering
    \includegraphics[width=0.45\textwidth]{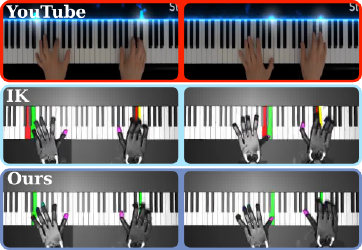}
    \caption{Qualitative comparison of hand poses. Top: Youtube video, Middle: \gls{ik} solution given the video. Bottom: After residual \gls{rl}.}
    \label{fig:youtube_qualitative}
\end{wrapfigure}
\\
\textbf{Inverse Kinematics (IK)~\cite{pink2024}} Given a demonstration fingertip trajectory $\tau_{\vx}$, a Quadratic Programming-based \gls{ik} solver~\cite{pink2024} is used to compute a target joint position trajectory and execute it open-loop. 

We select 10 clips with diverse levels of difficulty from the collected dataset. We individually train specialized policies for each of the 10 clips using both the baseline and our methods. Subsequently, we assess and compare their performance based on the achieved F1 Score.
\\
\textbf{Performance.} As shown in \Cref{fig:single}, our method consistently outperforms the Robopianist baseline for all 10 clips, achieving an average F1 score of 0.94 compared to the baseline's 0.75. We attribute this improvement to the incorporation of human priors, which narrows the RL search space to a favorable subspace, thereby encouraging the algorithm to converge towards more optimal policies. Additionally, the \gls{ik} method achieves an average F1 score of 0.72, only slightly lower than the baseline. This demonstrates the effectiveness of incorporating human priors, providing a strong starting point for \gls{rl}. 
\\
\textbf{Impact of Elements.} Our \gls{rl} method incorporates two main elements: Style-mimicking reward and Residual learning.
We individually exclude each element to investigate their respective influences on policy performance (See \Cref{fig:single}).
We clearly observe the critical role of residual learning implying the benefit of exploiting human demonstrations as nominal behavior.
We observe a marginal performance increase of 0.03 when excluding the style-mimicking reward; however, this also results in a larger discrepancy between the fingertip trajectory of the robot and the human.
Thus, the weight of the style-mimicking reward can be viewed as a parameter that controls the human likeness of the learned robot actions.
\\
\textbf{Qualitative comparison.} We present a qualitative comparison of the human likeness of the robot motion in \Cref{fig:youtube_qualitative} and the attached videos.
We inspect the hand poses for certain frames and observe that the \gls{ik} nominal behavior leads the robot to place the fingers in positions similar to those in Youtube videos. The \gls{rl} policy then slightly adapts the fingertip positions to press the keys correctly.








\subsection{Evaluation of model design strategies for policy distillation}
This section focuses on the evaluation of policy distillation for playing different songs. 
We evaluate the influence of different policy design strategies on the agent's performance. 
We aim to assess (\textbf{1}) the impact of integrating a pre-trained observation encoder to induce spatially consistent features, (\textbf{2}) the impact of a hierarchical design of the policy, and (\textbf{3}) the performance of different generative models on piano-playing data.

We propose two base policies, \textbf{Two-stage Diff} and \textbf{Two-stage Diff-res} policy. Both of them utilize hierarchical policies and goal representation learning, as described in \Cref{sec: dis}. The only difference between them is: the low-level policy of \textbf{Two-stage Diff} directly predicts the target joints, while \textbf{Two-stage Diff-res} predicts the residual term of an \gls{ik} solver.
Both high- and low-level policies are trained with Denoising Diffusion Probabilistic Models (DDPM) \cite{ho2020denoising}. 
The high-level policy is trained to predict the fingertip trajectory for 4 timesteps given the SDF embedding (See \Cref{app:bc}) of goals over 10 timesteps, while the low-level policy predicts the robot actions or residuals for 4 timesteps given the fingertip trajectory. Note that the entire dataset is used for training the high-level policy, while only around 40 \% of the collected clips (110K state-action pairs) are trained with \gls{rl} and further used for training the low-level policy.
The detailed network implementation is described in \Cref{app:network}.

Then, to analyze the impact of each variable, we design four variants of the Two-stage Diffusion policy.
To evaluate (\textbf{1}) the impact of integrating a pre-trained observation encoder, we train a model without the SDF embedding representation for the goal (\textbf{w/o SDF}).
To evaluate (\textbf{2}) the impact of the hierarchical architecture, we train a \textbf{One-stage} Diffusion policy that directly predict the joint space actions given the goal.
Finally, to evaluate (\textbf{3}) the influence of using different generative models, we train a Two-stage \textbf{BeT}, that replaces Diffusion models with Behavior-Transformers~\cite{shafiullah2022behavior}.
We also consider as baselines a \textbf{Multi-task RL} policy and a \textbf{BC} policy with \textbf{MSE} Loss. We provide further details of the models in \Cref{app:exp2}.

\textbf{Results} As shown in \Cref{tab:multi_datasets}, despite that Multi-task RL has the highest precision on the test dataset (this is because it barely presses any keys), our methods (Two-stage Diff and Two-stage Diff-res) outperform the others in all metrics on both training and test datasets. We also observe that the incorporation of SDF embedding for goal representation leads to better performance, especially on the test dataset, which demonstrates the impact of goal representation on policy generalization. Furthermore, we observe a slight performance enhancement when the model predicts the residual term of \gls{ik} (Two-stage Diff-res). We speculate that this improvement stems from our training data being generated through residual RL, which leverages the output of the IK solver as a prior. This approach likely causes the learned actions to correlate with the outputs of the IK solver.

\begin{table}
\centering
\begin{tblr}{
    colspec={lc| cc | cc | c | c | c},
    column{1-2}={font=\footnotesize},
    column{2-70}={c,font=\footnotesize}
}
    \SetCell[c=2,r=1]{l} {} &  & Multi-RL & BC-MSE & \textbf{Two-Stage Diff} &\textbf{-res} & w/o SDF & One-Stage & BeT\\
    \hline
    \SetCell[c=1,r=3]{l} \rotatebox{90}{\textbf{Train}} & P &  0.85 & 0.56 & 0.87 &\textbf{0.89} & 0.86  & 0.53 & 0.63 \\
    & R & 0.20 & 0.29 & 0.78  & \textbf{0.80} & 0.76 & 0.34  & 0.42 \\
    &  F1 & 0.12  & 0.30 & 0.81 & \textbf{0.82} & 0.78 & 0.35 & 0.49 \\
    \hline
    \SetCell[c=1,r=3]{l} \rotatebox{90}{\textbf{Test}} & P & \textbf{0.95} & 0.54 & 0.69 & 0.71 & 0.66 & 0.58 & 0.53 \\
    & R & 0.18  & 0.22 & 0.54 & \textbf{0.55} & 0.49 & 0.27 & 0.30 \\
    &  F1 & 0.13 & 0.21 & 0.56 & \textbf{0.57} & 0.51 & 0.26 & 0.31\\
    \hline
\end{tblr}
\caption{Quantitative results evaluated on Training and Test Datasets. Test datasets consist of 12 clips unseen in the training dataset. We report Precision (P), Recall (R) and F1-score (F1).}
\label{tab:multi_datasets}
\end{table}








\begin{figure}[htbp]
  \centering
  \includegraphics[width=\textwidth]{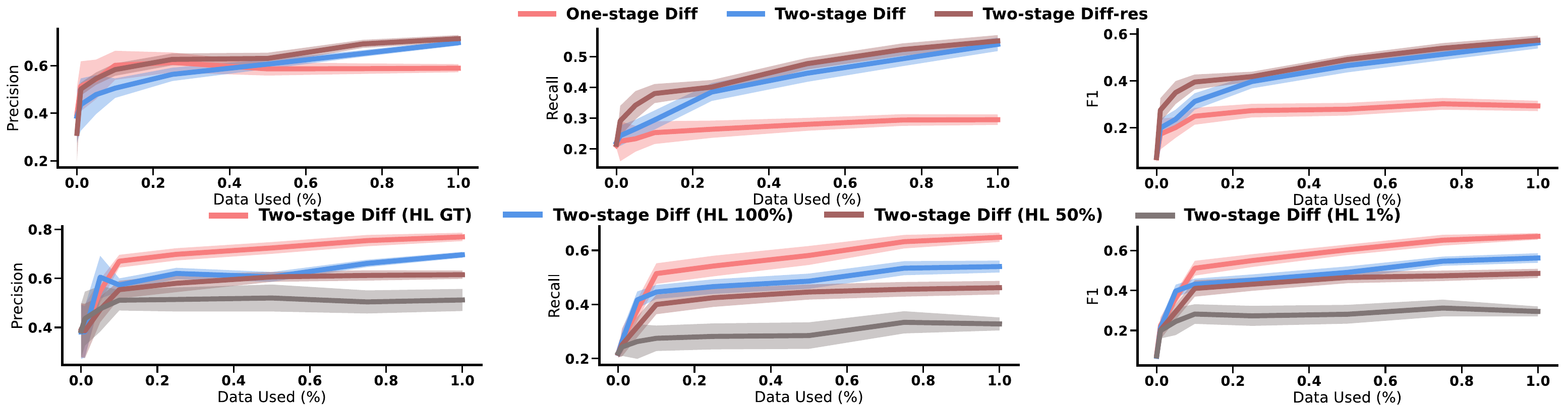}
  \caption{Precision and Recall for three different policy architectures trained with varying amount of data volumes evaluated on the test dataset. \textbf{Top}: Models are trained with the same proportion of high-level and low-level datasets. \textbf{Bottom}: Models are trained with different proportions of high-level and low-level datasets. The x-axis represents the percentage of the low-level dataset utilized, while HL \% indicates the percentage of the high-level dataset used.}
  \label{fig:perf_vs_data}
\end{figure}

\subsection{Evaluations on the impact of the data in the generalization}

In this section, we investigate the impact of scaling training data on the generalization capabilities of the agent.
We evaluate three policy designs (\textbf{One-stage Diff}, \textbf{Two-stage Diff}, and \textbf{Two-stage Diff-res}). We train them using various proportions of the dataset, and evaluate their performance on the test dataset (see \Cref{fig:perf_vs_data} Top). Note that One-stage Diff uses the same dataset as the low-level policy of Two-stage Diff.
\\
\textbf{Results.} We observe that both Two-stage Diff and Two-stage Diff-res show consistent performance improvement with increasing used data. This trend implies that the two-stage policies have not yet reached their performance saturation with the given data and could potentially continue to benefit from additional training data in future works. 

\textbf{Evaluation on imbalance training datasets.}
We further employ different combinations of the high-level and low-level policies of Two-stage Diff trained with different proportions of the dataset and assess their performance. In addition, we introduce an oracle high-level policy, which outputs the ground-truth fingertip position from human demonstration videos. The results (see \Cref{fig:perf_vs_data} Bottom) demonstrate that the overall performance of policy is significantly influenced by the quality of the high-level policy. Low-level policies paired with Oracle high-level policies consistently outperform the ones paired with other high-level policies. Besides, we observe early performance convergence with increasing training data when paired with a low-quality high-level policy. Specifically, with the HL $1\%$ policy and HL $50\%$, performance almost converged with around $10\%$ and $50\%$ low-level data, respectively.

\subsection{Limitations}
\textbf{Inference Speed}
One of the limitations is the inference speed. The models operate with an inference frequency of approximately 15Hz on an RTX 4090 machine, which is lower than the standard real-time demand on hardware. Future works can employ faster diffusion models, e.g., DDIM \cite{song2020denoising}, to speed up the inference.
\\
\textbf{Out-of-distribution Data}
Most of the songs in our collected dataset are of modern style. When evaluating the model on the dataset from \cite{zakka2023robopianist}, which mainly contains classical songs, the performance degrades. This discrepancy implies the model's limited generalization across songs of different styles. Future work can collect more diverse training data to improve this aspect.
\\
\textbf{Acoustic Experience}
Although the policy achieves up to 56\% F1-score on unseen songs, we found that higher accuracy is still necessary to make the song acoustically appealing and recognizable. Future work should focus on improving this accuracy to enhance the overall acoustic experience.


\section{Conclusion}
\label{sec:conclusion}
In this work, we present PianoMime, a framework for training a generalist robotic pianist using internet video sources. We start by training song-specific policies with residual RL, enabling the robot to master individual songs by mimicking human pianists.
Subsequently, we train a single behavioral cloning policy that mimics these song-specific policies to play unseen songs. The policy leverages three key techniques: goal representation learning, policy hierarchy, and expressive generative models. The resulting policy demonstrates an impressive generalization capability, achieving an average F1-score of 70\% on unseen songs. This also highlights that leveraging internet data can be highly useful for training generalist robotic agents.

\clearpage
\acknowledgments{If a paper is accepted, the final camera-ready version will (and probably should) include acknowledgments. All acknowledgments go at the end of the paper, including thanks to reviewers who gave useful comments, to colleagues who contributed to the ideas, and to funding agencies and corporate sponsors that provided financial support.}


\bibliography{bibliography}  

\newpage
\appendix
\section{Retargeting: From human hand to robot hand}
\label{app:retarget}
\begin{figure}
    \centering
    \includegraphics[width=\textwidth]{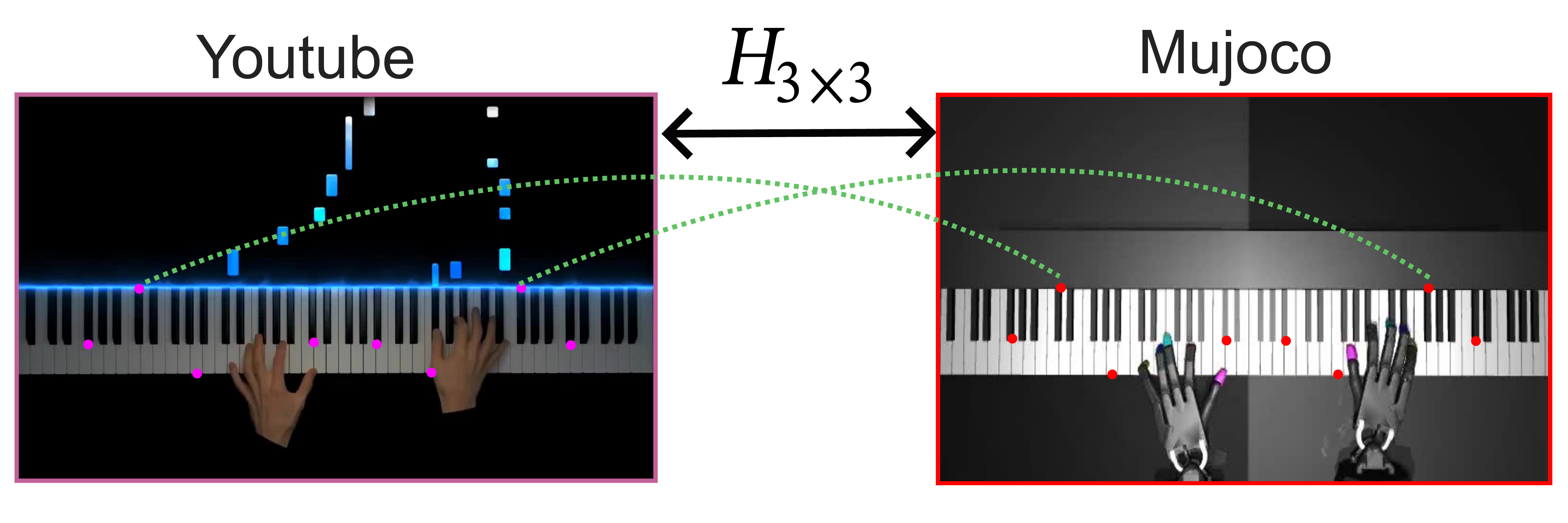}
    \caption{Compute homography matrix given 8 correspondence feature points.}
    \label{fig:piano}
\end{figure}

To retarget from the human hand to robot hand, we follow a structured process. 
\\
\textbf{Step 1: Homography Matrix Computation} Given a top-view piano demonstration video, we firstly choose $n$ different feature points on the piano. These points could be center points of specific keys, edges, or other identifiable parts of the keys that are easily recognizable (See \Cref{fig:piano}). Due to the uniform design of pianos, these points represent the same physical positions in both the video and Mujoco. Given the chosen points, we follow the Eight-point Algorithm to compute the Homography Matrix $H$ that transforms the pixel coordinate in videos to the x-y coordinate in Mujoco (z-axis is the vertical axis). 
\\\\
\textbf{Step 2: Transformation of Fingertip Trajectory} We then obtain the human fingertip trajectory with MediaPipe \cite{lugaresi2019mediapipe}. We collect the fingertips positions every 0.05 seconds. Then we transform the human fingertip trajectory within pixel coordinate into the Mujoco x-y 2D coordinate using the computed homography matrix $H$. 
\\\\
\textbf{Step 3: Heuristic Adjustment for Physical Alignment} We found that the transformed fingertip trajectory might not physically align with the notes, which means there might be no detected fingertip that physically locates at the keys to be pressed or the detected fingertip might locate at the border of the key (normally human presses the middle point of the horizontal axis of the key). This misalignment could be due to the inaccuracy of the hand-tracking algorithm and the homography matrix. Therefore, we perform a simple heuristic adjustment on the trajectory to improve the physical alignment. Specifically, at each timestep of the video, we check whether there is any fingertip that physically locates at the key to be pressed. If there is, we adjust its y-axis value to the middle point of the corresponding key. Otherwise, we search within a small range, specifically the neighboring two keys, to find the nearest fingertip. If no fingertip is found in the range or the found fingertip has been assigned to another key to be pressed, we then leave it. Otherwise, we adjust its y-axis value to the center of the corresponding key to ensure proper physical alignment. 
\\\\
\textbf{Step 4: Z-axis Value Assignment} Lastly, we assign the z-axis value for the fingertips. For the fingertips that press keys, we set their z-axis values to $0$. For other fingertips, we set their z-axis value to $2 \cdot h_{key}$, where $h_{key}$ is the height of the keys in Mujoco. 

\section{Implementation of Inverse Kinematics Solver}
The implementation of the IK solver is based on the approach of \cite{pink2024}. The solver addresses multiple tasks simultaneously by formulating an optimization problem and find the optimal joint velocities that minimize the objective function. The optimization problem is given by:
\begin{equation}
\min_{\dot{q}} \sum_i w_i \| J_i \dot{q} - K_i v_i \|^2,
\end{equation}
where $w_i$ is the weight of each task, $K_i$ is the proportional gain and $v_i$ is the velocity residual.
We define a set of 10 tasks, each specifying the desired position of one of the robot fingertips. We do not specify the desired quaternions. All the weights $w_i$ are set to be equal. We use quadprog \footnote{https://github.com/quadprog/quadprog} to solve the optimization problem with quadratic programming. The other parameters are listed in \Cref{table:IK}.
\begin{table}[h]
  \caption{The parameters of IK solver}
  \centering
  \begin{tblr}{
      colspec={lllll},
      row{1}={font=\bfseries},
      row{even}={bg=gray!10},
    }
    \textbf{Parameter}  & Value  \\
    \hline
    Gain & 1.0 \\
    Limit Gain & 0.05 \\
    Damping & 1e-6 \\
    Levenberg-Marquardt Damping & 1e-6 

  \end{tblr}
\label{table:IK}
\end{table}

\section{Detailed MDP Formulation of Song-specific Policy}
\label{app:mdp}
\begin{table}[h]

  \caption{The detailed reward function to train the song-specific policy. The Key Press reward is the same as in \cite{zakka2023robopianist}, where $k_s$ and $k_g$
  represent the current and the goal states of the key respectively, and g is a function that transforms the distances
  to rewards in the [0, 1] range. $p_{df}$ and $p_{rf}$ represent the fingertip positions of human demonstrator and robot respectively.}
  \centering
  \begin{tblr}{
      colspec={lllll},
      row{1}={font=\bfseries},
      row{even}={bg=gray!10},
    }
    \textbf{Reward}         & Formula  & Weight  & Explanation  \\
    \hline
    Key Press & $0.5 \cdot g(\|k_s - k_g\|_2) + 0.5 \cdot (1 - \mathbf{1}_{\text{false positive}})$ & 2/3 & Press the right keys and \newline only the right keys \\
    Mimic & $g(\|p_{df} - p_{rf}\|_2)$ & 1/3 & Mimic the demonstrator's \newline fingertip trajectory
  \end{tblr}
\end{table}

\begin{table}[h]
  \caption{The observation space of song-specific agent.}
  \centering
  \begin{tblr}{
      colspec={lllll},
      row{1}={font=\bfseries},
      row{even}={bg=gray!10},
    }
    \textbf{Observation}         & Unit  & Size  \\
    \hline
    Hand and Forearm Joint Positions & Rad & 52 \\
    Hand and forearm Joint Velocities & Rad/s & 52 \\
    Piano Key Joint Positions & Rad & 88 \\
    Piano key Goal State & Discrete & 88 \\
    Demonstrator Forearm and Fingertips Cartesian Positions & m & 36 \\
    Prior control input $\Tilde{u}$ (solved by IK) & Rad & 52 \\
    Sustain Pedal state & Discrete & 1
  \end{tblr}
\end{table}

\begin{table}[h]
  \caption{The action space of song-specific agent.}
  \centering
  \begin{tblr}{
      colspec={lllll},
      row{1}={font=\bfseries},
      row{even}={bg=gray!10},
    }
    \textbf{Action}         & Unit  & Size  \\
    \hline
    Target Joint Positions & Rad & 46 \\
    Sustain Pedal & Discrete & 1
  \end{tblr}
\end{table}
\section{Training Details of Song-specific Policy}
We use PPO \cite{schulman2017proximal} (implemented by StableBaseline 3 \cite{stable-baselines3}) to train the song-specific policy with residual RL(See Algorithm \ref{alg:single}). All of the experiments are conducted using the same network architecture and tested using 3 different seeds. Both actor and critic networks are of the same architecture, containing 2 MLP hidden layers with 1024 and 256 nodes, respectively, and GELU \cite{hendrycks2016gaussian} as activation functions. The detailed hyperparameters of the networks are listed in Table \ref{table:ppo_hyperparameters}.

\begin{table}[h]
  \caption{The Hyperparameters of PPO}
  \centering
  \begin{tblr}{
      colspec={lllll},
      row{1}={font=\bfseries},
      row{even}={bg=gray!10},
    }
    \textbf{Hyperparameter}  & Value  \\
    \hline
    Initial Learning Rate & 3e-4 \\
    Learning Rate Scheduler & Exponential Decay \\
    Decay Rate & 0.999 \\
    Actor Hidden Units & 1024, 256 \\
    Actor Activation & GELU \\
    Critic Hidden Units & 1024, 256 \\
    Critic Activation & GELU \\
    Discount Factor & 0.99 \\
    Steps per Update & 8192 \\
    GAE Lambda & 0.95 \\
    Entropy Coefficient & 0.0 \\
    Maximum Gradient Norm & 0.5 \\
    Batch Size & 1024 \\
    Number of Epochs per Iteration & 10 \\
    Clip Range & 0.2 \\
    Number of Iterations & 2000 \\
    Optimizer & Adam 
  \end{tblr}
\label{table:ppo_hyperparameters}
\end{table}

\begin{algorithm}[h]
\caption{Training of the song-specific policy with residual RL}
\label{alg:single}
\begin{algorithmic}[1]
\STATE Initialize actor network $\pi_\theta$
\STATE Initialize critic network $v_\phi$

\FOR{$i = 1 : N_{iteration}$}
    \STATE \COMMENT{Collect trajectories}
    \FOR{$t = 1:T$}
        \STATE Get human demonstrator fingertip position $x_t$ and observation $o_t$
        \STATE Compute the prior control signal that tracks $x_t$ with the IK controller $\Tilde{u}_t = ik(x_t, o_t)$
        \STATE Run policy to get the residual term $r_t = \pi_\theta(o_t)$
        \STATE Compute the adapted control signal $u_t = \Tilde{u_t}+r_t$
        \STATE Execute $u_t$ in environment and collect {$s_t, u_t, r_t, s_{t+1}$}
    \ENDFOR
    \STATE \COMMENT{Update networks}
    \FOR{$n = 1: N$}
        \STATE Sample a batch of transitions $\{(s_j, u_j, r_j, s_{j+1})\}$ from the collected trajectories 

        \STATE Update the actor and critic network with PPO
    \ENDFOR
\ENDFOR
\end{algorithmic}
\end{algorithm}

\section{Representation Learning of Goal}
\begin{figure}
    \centering
    \includegraphics[width=\textwidth]{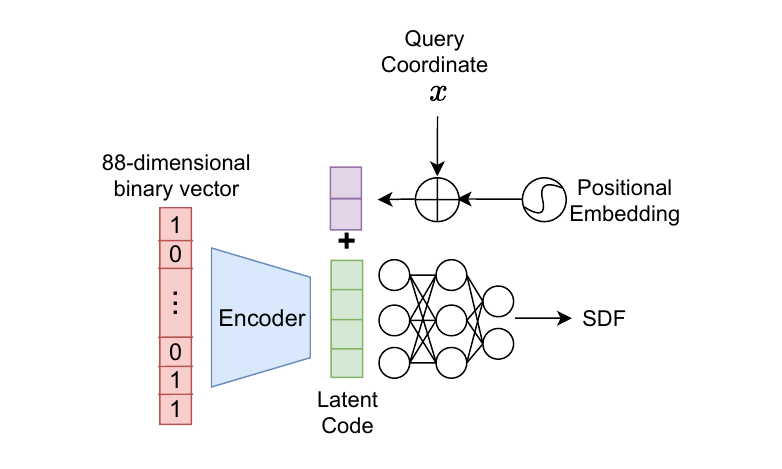}
    \caption{1) Encoding: The encoder compresses the binary representation of the goal into latent code. 2) Decoding: A 3D query coordinate $x$ is randomly sampled. A neural network predicts the SDF value given the positional encoding of $x$ and the latent code.}
    \label{fig:sdf}
\end{figure}

We train an autoencoder to learn a geometrically continuous representation of the goal (See Figure \ref{fig:sdf} and Algorithm \ref{alg:sdf}). During the training phase, the encoder $\mathcal{E}$, encodes the original 88-dimensional binary representation of a goal piano state $\musEighth_t$ into a 16-dimensional latent code $z$. The positional encoding of a randomly sampled 3D query coordinate $x$ is then concatenated with the latent code $z$ and passed through the decoder $\mathcal{D}$. We use positional encoding here to represent the query coordinate more expressively. The decoder is trained to predict the SDF $f(x, \musEighth_t)$. We define the SDF value of $x$ with respect to $\musEighth_t$ as the Euclidean distance between the $x$ and the nearest key that is supposed to be pressed in $\musEighth_t$, mathematically expressed as: 
\begin{equation}
    \text{SDF}(x, \musEighth_t) = \min_{p \in \{p_i \mid \musEighth_{t,i} = 1\}} \| x - p \|,
\end{equation} where $pi$ represents the position of the 
$i$-th key on the piano.  The encoder and decoder are jointly optimized to minimize the reconstruction loss:
\begin{equation}
    L(x, , \musEighth_t)=(\text{SDF}(x, \musEighth_t) - \mathcal{D}(\mathcal{E}(v, x)))^2.
\end{equation}
\\
We pre-train the autoencoder using the \textbf{GiantMIDI} dataset \footnote{https://github.com/bytedance/GiantMIDI-Piano}, which contains 10K piano MIDI files of 2,786 composers. The pre-trained encoder maps the $\musEighth_t$ into the 16-dimensional latent code, which serves as the latent goal for behavioral cloning. The encoder network is composed of four 1D-convolutional layers, followed by a linear layer. Each successive 1D-convolutional layer has an increasing number of filters, specifically 2, 4, 8, and 16 filters, respectively. All convolutional layers utilize a kernel size of 3. The linear layer transforms the flattened output from the convolutional layers into a 16-dimensional latent code. The decoder network is a MLP with 2 hidden layers, each with 16 neurons. We train the autoencoder for 100 epochs with a learning rate of $1e-3$.
\label{app:bc}

\begin{algorithm}[h]
\caption{Training of the goal autoencoder}
\label{alg:sdf}
\begin{algorithmic}[1]
\STATE Initialize encoder $\mathcal{E}_\mathcal{\phi}$
\STATE Initialize decoder $\mathcal{D}_\mathcal{\psi}$

\FOR{$i = 1 : N_{epoch}$}
    \FOR{$j = 1 : N_{batch}$}
        \FOR{each goal $v$ in batch}
            \STATE Compute the latent code $\mathbf{z} = \mathcal{E}_\mathcal{\psi}(\musEighth_t)$ 
            \STATE Sample a 3D coordinate as query $\mathbf{x} = \text{Sample3DCoordinate()}$
            \STATE Compute the positional encoding of query $\mathbf{pe} = \text{PositionalEncoding(}x)$
            \STATE Compute the output of the decoder conditioned by the query $\mathcal{D}_\mathcal{\phi}(z, pe)$  
            \STATE Compute the SDF value of query $\text{SDF}(x, \musEighth_t)$
            \STATE Compute the reconstruction loss $L$
        \ENDFOR
        \STATE Compute the sum of the loss                      
        \STATE Compute the gradient
        \STATE Update network parameter $\mathcal{\phi}, \mathcal{\psi}$
    \ENDFOR
\ENDFOR
\end{algorithmic}
\end{algorithm}

\section{Training Details of Diffusion Model}
\label{app:network}
All the diffusion models utilized in this work, including One-stage Diff, the high-level and low-level policies of Two-stage Diff, Two-stage Diff-res and Two-stage Diff w/o SDF, share the same network architecture. The network architecture are the same as the U-net diffusion policy in \cite{chi2023diffusion} and optimized with DDPM \cite{ho2020denoising}, except that we use temporal convolutional networks (TCNs) as the observation encoder, taking the concatenated goals (high-level policy) or fingertip positions (low-level policy) of several timesteps as input to extract the features on temporal dimension. Each level of U-net is then conditioned by the outputs of TCNs through FiLM \cite{perez2018film}. 
\\\\
High-level policies take the goals over 10 timesteps and the current fingertip position as input and predict the human fingertip positions. In addition, we add a standard gaussian noise on the current fingertip position during training to facilitate generalization. We further adjust the y-axis value of the fingertips pressing the keys in the predicted high-level trajectories to the midpoint of the keys. This adjustment ensures closer alignment with the data distribution of the training dataset. Low-level policies take the predicted fingertip positions and the goals over 4 timesteps, the proprioception state as input predict the robot actions. The proprioception state includes the robot joint positions and velocities, as well as the piano joint positions. We use 100 diffusion steps during training. To achieve high-quality results during inference, we find that at least 80 diffusion steps are required for high-level policies and 50 steps for low-level policies. 
\begin{table}[H]
  \caption{The Hyperparameters of DDPM}
  \centering
  \begin{tblr}{
      colspec={lllll},
      row{1}={font=\bfseries},
      row{even}={bg=gray!10},
    }
    \textbf{Hyperparameter}  & Value  \\
    \hline
    Initial Learning Rate & 1e-4 \\
    Learning Rate Scheduler & Cosine \\
    U-Net Filters Number & 256, 512, 1024 \\
    U-Net Kernel Size & 5 \\
    TCN Filters Number & 32, 64 \\
    TCN Kernel Size & 3 \\
    Diffusion Steps Number & 100 \\
    Batch Size & 256 \\
    Number of Iterations & 800 \\
    Optimizer & AdamW \\
    EMA Exponential Factor & 0.75 \\
    EMA Inverse Multiplicative Factor & 1 
  \end{tblr}
\label{table:ppo_hyperparameters}
\end{table}

\section{Policy Distillation Experiment}
\label{app:exp2}

\textbf{Two-stage Diff w/o SDF} We directly use the binary representation of goal instead of the SDF embedding representation to condition the high-level and low-level policies.
\\\\
\textbf{Two-stage Diff-res} We employ an IK solver to compute the target joints given the fingertip positions predicted by the high-level policy. The low-level policy predicts the residual terms of IK solver instead of the robot actions.
\\\\
\textbf{Two-stage BeT} We train both high-level and low-level policies with Behavior Transformer \cite{shafiullah2022behavior} instead of DDPM. The hyperparameter of Bet is listed in \Cref{table:bet}.
\\\\
\textbf{One-stage Diff} We train a single diffusion model to predict the robot actions given the SDF embedding representation of goals and the proprioception state.
\\\\
\textbf{Multi-task RL} We create a multi-task environment where for each episode a random song is sampled from the dataset. Consequently, we use Soft-Actor-Critic (SAC) \cite{haarnoja2018soft} to train a single agent within the environment. Both the actor and critic networks are MLPs, each with 3 hidden layers, and each hidden layer contains 256 neurons. The reward function is the same as that in \cite{zakka2023robopianist}.
\\\\
\textbf{BC-MSE} We train a feedforward network to predict the robot action of next timestep conditioned on the binary representation of goal and proprioception state with MSE loss. The feedforward network is a MLP with 3 hidden layers, each with 1024 neurons. 
\begin{table}[h]
  \caption{The Hyperparameters of Behavior Transformer}
  \centering
  \begin{tblr}{
      colspec={lllll},
      row{1}={font=\bfseries},
      row{even}={bg=gray!10},
    }
    \textbf{Hyperparameter}  & Value  \\
    \hline
    Initial Learning Rate & 3e-4 \\
    Learning Rate Scheduler & Cosine \\
    Number of Discretization Bins & 64 \\
    Number of Transformer Heads & 8 \\
    Number of Transformer Layers & 8 \\
    Embedding Dimension & 120\\
    Batch Size & 256 \\
    Number of Iterations & 1200 \\
    Optimizer & AdamW \\
    EMA Exponential Factor & 0.75 \\
    EMA Inverse Multiplicative Factor & 1 
  \end{tblr}
\label{table:bet}
\end{table}

\section{F1 Score of All Trained Song-Specific Policies}
\begin{figure}[h]
    \centering
    \rotatebox{270}{\includegraphics[width=1.5\textwidth]{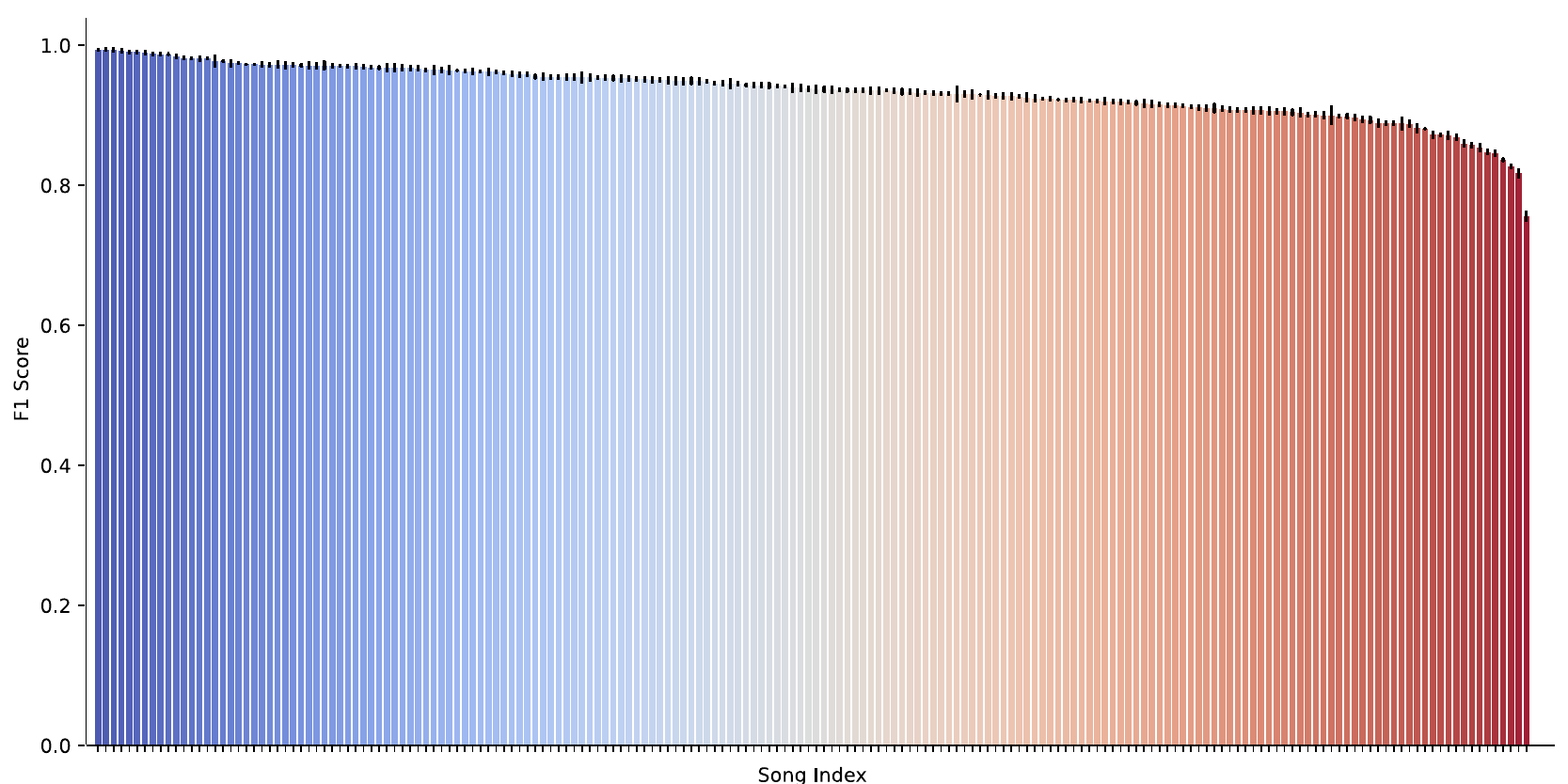}}
    \caption{F1 score of all 184 trained song-specific policies (descending order)}
    \label{fig:all_f1_scores}
\end{figure}

\Cref{fig:all_f1_scores} shows the F1 score of all song-specific policies we trained. 

\section{Detailed Results on Test Dataset}
In \Cref{tab:test} and \Cref{tab:etude}, we show the Precision, Recall and F1 score of each song in our collected test dataset and the Etude-12 dataset from \cite{zakka2023robopianist}, achieved by Two-stage Diff and Two-stage Diff-res, respectively. We observe an obvious performance degradation when testing on Etude-12 dataset. We suspect that the reason is due to out-of-distribution data, as the songs in the Etude-12 dataset are all classical, whereas our training and test dataset primarily consists of modern songs.
\begin{table}[h]
\caption{Quantitative results of each song in the our collected test dataset}
\centering
\begin{tblr}{
    colspec={lccc | ccc},
    row{1}={font=\bfseries\footnotesize, bg=gray!10},
    row{2}={font=\footnotesize},
    row{even}={bg=gray!10},
    width=\linewidth,
    column{1}={font=\footnotesize},
    hline{2} = {2-7}{solid}
}
    \SetCell[c=1,r=2]{l} \textbf{Song Name} & \SetCell[c=3]{c} Two-stage Diff & & & \SetCell[c=3]{c} Two-stage Diff-res & & \\
    & Precision & Recall & F1 & Precision & Recall & F1 \\
    \hline
    Forester & 0.81 & 0.70 & 0.68 & 0.79 & 0.71 & 0.67 \\
    Wednesday  & 0.66 & 0.57 & 0.58 & 0.67 & 0.54 & 0.55 \\
    Alone    & 0.80 & 0.62 & 0.66 & 0.83 & 0.65 & 0.67 \\
    Somewhere Only We Know & 0.63 & 0.53 & 0.58 & 0.67 & 0.57 & 0.59 \\
    Eyes Closed & 0.60 & 0.52 & 0.53 & 0.61 & 0.45 & 0.50 \\
    Pedro       & 0.70 & 0.58 & 0.60 & 0.67 & 0.56 & 0.47 \\
    Ohne Dich   & 0.73 & 0.55 & 0.58 & 0.75 & 0.56 & 0.62 \\
    Paradise    & 0.66 & 0.42 & 0.43 & 0.68 & 0.45 & 0.47 \\
    Hope       & 0.74 & 0.55 & 0.57 & 0.76 & 0.58 & 0.62 \\
    No Time To Die  & 0.77 & 0.53 & 0.55 & 0.79 & 0.57 & 0.60 \\
    The Spectre   & 0.64 & 0.52 & 0.54 & 0.67 & 0.50 & 0.52 \\
    Numb   & 0.55 & 0.44 & 0.45 & 0.57 & 0.47 & 0.48 \\
    \textbf{Mean} & \textbf{0.69} & \textbf{0.54} & \textbf{0.56} & \textbf{0.71} & \textbf{0.55} & \textbf{0.57}
    
\end{tblr}
\label{tab:test}
\end{table}

\begin{table}[H]
\caption{Quantitative results of each song in the Etude-12 dataset}
\centering
\begin{tblr}{
    colspec={lccc | ccc},
    row{1}={font=\bfseries\footnotesize, bg=gray!10},
    row{2}={font=\footnotesize},
    row{even}={bg=gray!10},
    width=\linewidth,
    column{1}={font=\footnotesize},
    hline{2} = {2-7}{solid}
}
    \SetCell[c=1,r=2]{l} \textbf{Song Name} & \SetCell[c=3]{c} Two-stage Diff & & & \SetCell[c=3]{c} Two-stage Diff-res & & \\
    & Precision & Recall & F1 & Precision & Recall & F1 \\
    \hline
    FrenchSuiteNo1Allemande & 0.45 & 0.31 & 0.34 & 0.39 & 0.27 & 0.30 \\
    FrenchSuiteNo5Sarabande  & 0.29 & 0.23 & 0.24 & 0.24 & 0.18 & 0.19 \\
    PianoSonataD8451StMov    & 0.58 & 0.52 & 0.52 & 0.60 & 0.50 & 0.51 \\
    PartitaNo26   & 0.35 & 0.22 & 0.24 & 0.40 & 0.24 & 0.26 \\
    WaltzOp64No1 & 0.44 & 0.31 & 0.33 & 0.43 & 0.28 & 0.31 \\
    BagatelleOp3No4       & 0.45 & 0.30 & 0.33 & 0.45 & 0.28 & 0.32 \\
    KreislerianaOp16No8   &0.43 & 0.34 & 0.36 & 0.49 & 0.34 & 0.36 \\
    FrenchSuiteNo5Gavotte       & 0.34 & 0.29 & 0.33 & 0.41 & 0.31 & 0.33 \\
    PianoSonataNo232NdMov       & 0.35 & 0.24 & 0.25 & 0.29 & 0.19 & 0.21 \\
    GolliwoggsCakewalk       & 0.60 & 0.43 & 0.45 & 0.57 & 0.40 & 0.42 \\
    PianoSonataNo21StMov       & 0.32 & 0.22 & 0.25 & 0.36 & 0.23 & 0.25 \\
    PianoSonataK279InCMajor1StMov   & 0.43 & 0.35 & 0.35 & 0.53 & 0.38 & 0.39 \\
    \textbf{Mean} & \textbf{0.42} & \textbf{0.31} & \textbf{0.33} & \textbf{0.43} & \textbf{0.30} & \textbf{0.32}
    
\end{tblr}
\label{tab:etude}
\end{table}

\end{document}